\documentclass[conference]{IEEEtran}
\IEEEoverridecommandlockouts
\usepackage{amsmath,amssymb,amsfonts}
\usepackage{algorithmic}
\usepackage{graphicx}
\usepackage{textcomp}
\usepackage{xcolor}

\usepackage[%
    style=ieee,
    natbib=true
]{biblatex}
\bibliography{amoc21-darknet-discourse-influence-lit}

\usepackage{booktabs}

\DeclareGraphicsExtensions{.pdf,.ai,.jpg,.png}
\setkeys{Gin}{pagebox=artbox}
\graphicspath{{./amoc21-darknet-discourse-influence-figures/}}

\begin{document}

\title{Tracking Discourse Influence in Darknet Forums \\[.9ex]
\LARGE{Submission of Team SamSepi0l to the AMoC Hackathon}}

\author{
\IEEEauthorblockN{Christopher Akiki, Lukas Gienapp, and Martin Potthast}
\IEEEauthorblockA{\textit{Text Mining and Retrieval Group} \\
\textit{Leipzig University}}
}

\maketitle

\begin{abstract}
This technical report documents our efforts in addressing the tasks set forth by the 2021 AMoC (Advanced Modelling of Cyber Criminal Careers) Hackathon. Our main contribution is a joint visualisation of semantic and temporal features, generating insight into the supplied data on darknet cybercrime through the aspects of novelty, transience, and resonance, which describe the potential impact a message might have on the overall discourse in darknet communities. All code and data produced by us as part of this hackathon is publicly available.%
\footnote{https://github.com/webis-de/AMOC-21}
\end{abstract}

\section{Introduction}

The hackathon encompassed two separate tasks. The goal of the first task was to create an innovative approach to visualising the temporal nature of the dataset to allow for a longitudinal analysis of how significant events might affect the nature of messages exchanged on the forums. The second task seeks to perform authorship attribution to re-identify individuals' accounts across different forums. 
 
Both tasks aim at gaining novel insight into financially-motivated cybercrime on darknet markets. The hackathon makes use of a subset of two datasets: the Darknet Market Archives~\cite{gwern:2015} and the hacker forums of AZsecure.org~\cite{alsayra:2011}. The final dataset consists of 40~fora of the dark web. These fora typically serve as escrow spaces where buyers and sellers of illicit goods and services converge to conduct transactions.

\begin{figure*}
\centering
\includegraphics[width=\textwidth]{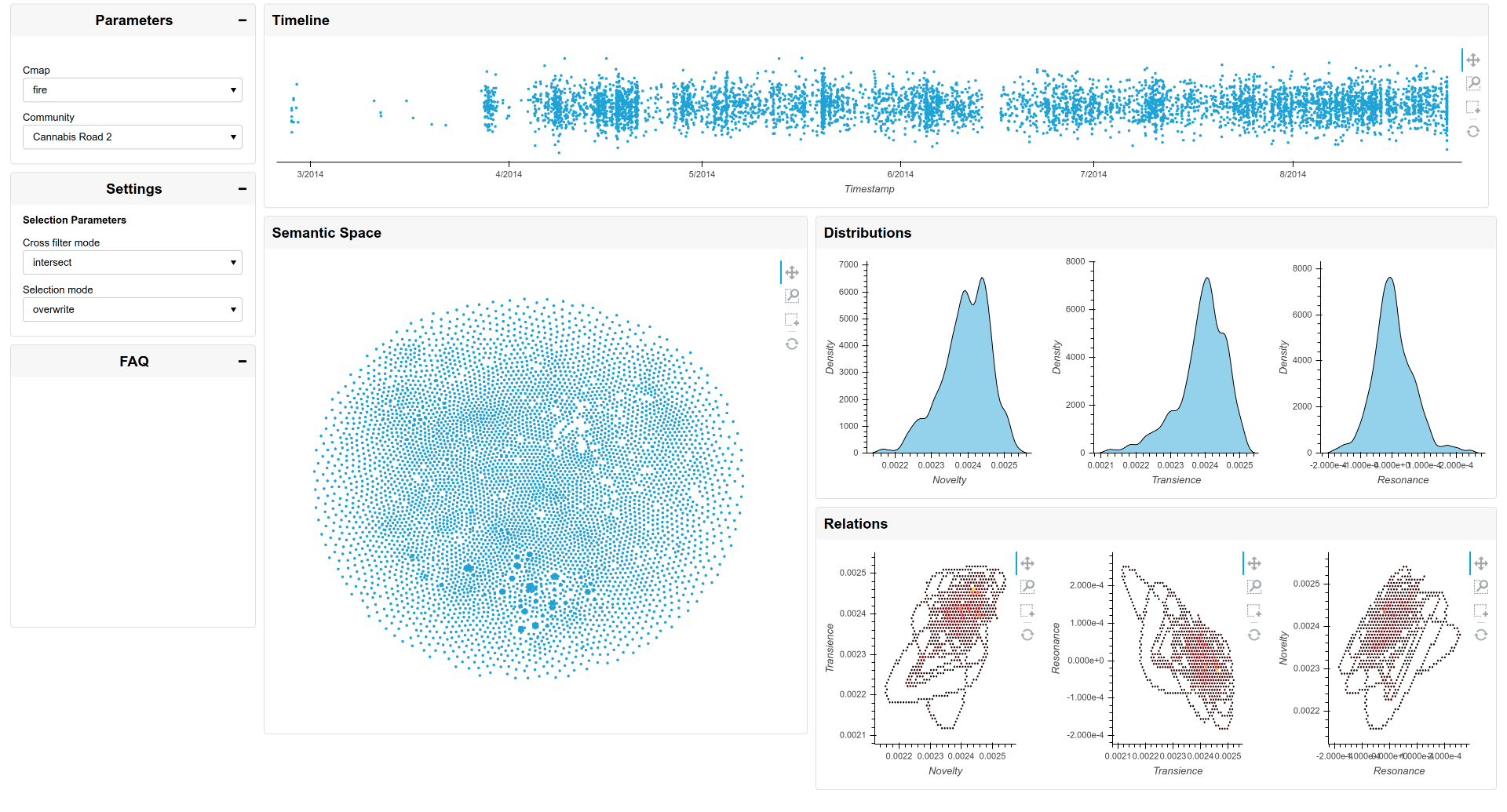}
\caption{Final Visualisation Dashboard}
\label{fig:final}
\end{figure*}

\begin{figure*}
\centering
\includegraphics[width=\textwidth]{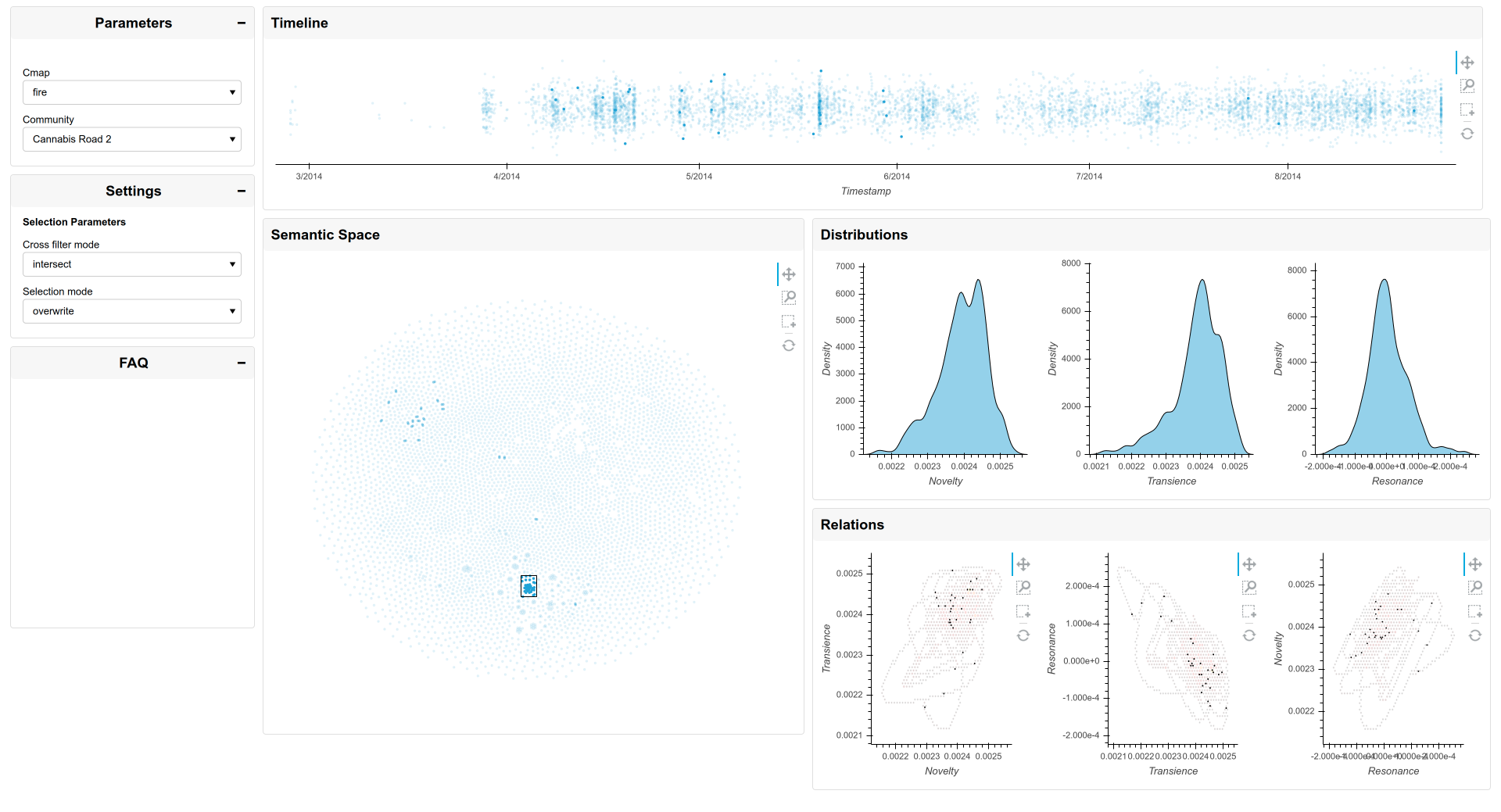}
\caption{Final Visualisation Dashboard with active selection}
\label{fig:final_selected}
\end{figure*}

\section{Methodological Approach}

This section details the methodological approaches and design decisions influencing our solutions to both of proposed tasks. 

\subsection{First Task}

The underlying goal of the visualisation approach we chose for Task~1 is to make the relation between the content of messages posted on dark web forums and the time messages were posted there both visible and explorable to an end user. Therefore, the visualisation dashboard we created (see Figure~\ref{fig:final}) includes three distinct modes of visualisation. The first one is the temporal nature, represented by a simple timeline at the top, allowing users to browse the data by making time-based selections. The second is content, represented by the semantic space embedding to the left of the visualisations; here, posts are plotted by their position in the semantic space of their respective community. The third component visualises the interaction effect between time and semantic space, plotting the three features novelty, transience, and resonance. For all three, we plot both the distribution, as well as the x-y~interaction plot between them.

To calculate the position of messages in their communities' semantic space, we rely on the transformer-encoder-based variant of the Universal Sentence Encoder (USE)~\cite{cer:2018} to calculate phrase-level embeddings for the body text of all posts. The USE model consists of a transformer-encoder architecture very similar to BERT~\cite{devlin:2019}, albeit trained with two key differences: first through the use of the rule-based PBT tokenizer, and second through a more downstream-aware multi-task supervised pretraining regime.

The resulting vector space spanned by the 512-dimensional embeddings USE produces can be used to calculate the semantic similarity of texts. To make these high-dimensional semantic relations visible to the end user, we resort to manifold learning whereby we try to learn a 2-dimensional non-linear topological space that best approximates the data in low-dimensions. To that end, we experimented both with t-distributed Stochastic Neighbor Embedding (t-SNE)~\cite{hinton:2002} and Uniform Manifold Approximation and Projection (UMAP)~\cite{mcinnes:2018,nolet:2020}, and ultimately chose t-SNE as it provided for a better visual result upon manual inspection. Manifold learning was performed separately per community as the final visualisation is centred around community-specific views.

Furthermore, we performed density-based clustering using the DBSCAN~\citep{ester:1996} algorithm to make different semantic groupings in the data more easily visible. 

To estimate the interaction effect between time and semantic space, we expand upon the methods developed by~\citet{barron:2018}, originally meant to study a political body---that of the national assembly of revolutionary France---as a heteroglossic system that evolves through time within a bounded political context. We find the parallel between a longitudinal corpus of political speeches and a longitudinal corpus of forum posts structurally similar enough to warrant an adaptation of their methods. This approach boils down to computing three longitudinal vectors using a sliding window approach: novelty, which is quantified by the divergence of a document to its local past; transience, which is quantified by the divergence of a document to its local future; and resonance, which quantifies the difference of these two dynamic quantities, measuring their interplay. We calculate the three features novelty, transience and resonance to model the influence to the communities discourse a single message has. 

In the context of this task, a high novelty value could, for example, be used to identify messages that introduce a new product to a darknet market, while an additional low transience value might help identifying members that are highly influential on the overall discourse of platform, and are key community leaders. However, we refrain from making any further assumptions in this direction because we lack the domain-specific knowledge required to make a useful interpretation of the collected data.

The features are calculated in a sliding-window manner: here, for each post~$p_t$ at point in time~$t$, its novelty is measured as the Kullback-Leibler divergence of a semantic probability distribution over~$p_t$ to the average distribution of all previous posts $p_{t-1}, ..., p_{t-n}$ in window of size~$n$. For transience, the same method is applied, but comparing to all following posts $p_{t+1}, ..., p_{t+n}$. Finally, resonance is measured as the asymmetric difference between novelty and transience. We infer a semantic probability distribution for each post~$p$ by applying a softmax function to the semantic embedding vector as produced by the USE (see previous section).

\subsection{Second Task}

We set about solving this task using the novel approach introduced by~\citet{sun:2020} and went as far as implementing it using TensorFlow and the Huggingface library~\cite{wolf:2020}. This approach leverages the generation dynamics of causal language models, GPT-2~\cite{radford:2019} in this instance, to compute a fingerprint for a given text. Upon deeper examination, it became clear to us that this method of fingerprinting text would be better suited for a side-channel scenario where one does not have access to the original text, but merely to the smart device upon which said text is generated.

As a fallback implementation, we started to apply the unmasking algorithm originally developed by~\citet{koppel:2004} and refined for the domain of short texts by~\citet{stein:2019h}. However, due to the short time frame of the hackathon and the time ``lost'' on the first approach, we did not finish this part of the task and can therefore not present meaningful results.

\section{Results}

\begin{figure*}%
\centering%
\includegraphics[width=.495\textwidth]{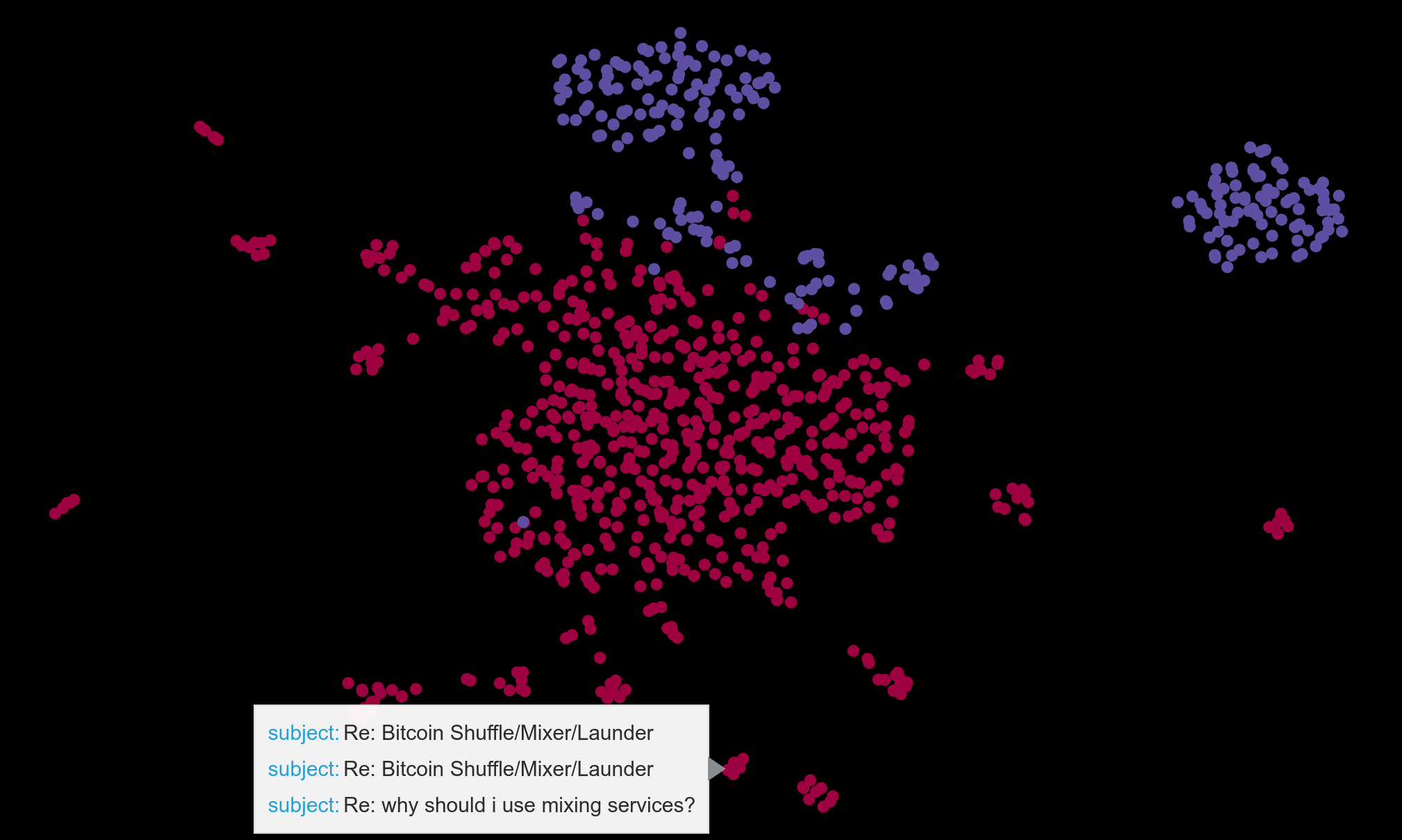}%
\hfill%
\includegraphics[width=.495\textwidth]{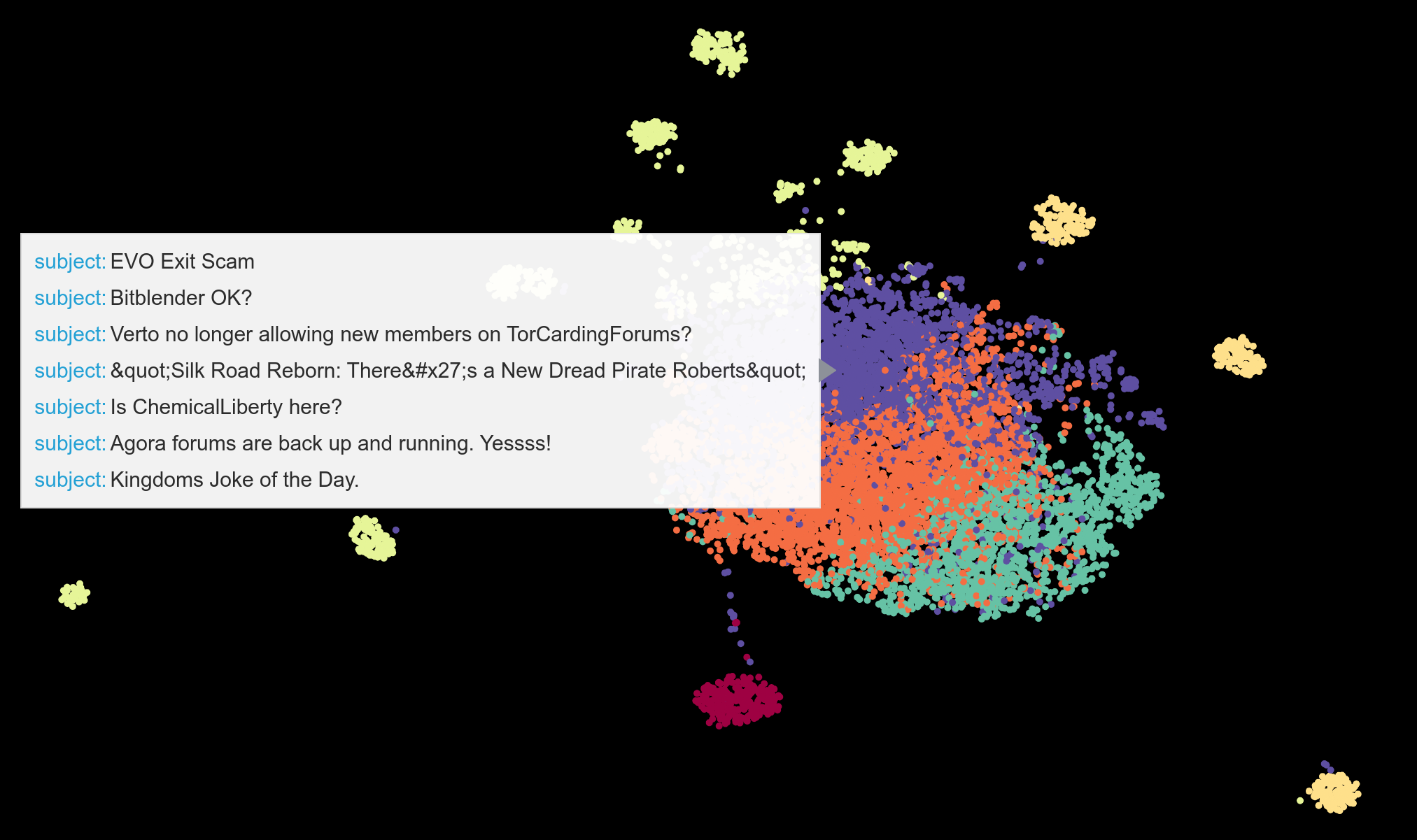}%
\caption{Example plot for semantic space with clusters hightlighted as produced by UMAP for the Hydra Forums (left) and for the Kingdom Forums (right).}%
\label{fig:individual}%
\end{figure*}

Our main contribution to the hackathon is the visualisation dashboard pictured in Figure~\ref{fig:final}. While we initially planned to include cluster information of semantic space, the clustering results are not displayed on the final visualisation as computations did not finish within time. However, cluster information is available in the individual visualisations as produced by UMAP (Figure~\ref{fig:individual}).

Furthermore, we implement an interactive component, such that if a user highlights a data point, or area of data in one of the plots, the corresponding data in other plots is highlighted as well (Figure~\ref{fig:final_selected}). 

Results are displayed per individual community. In the demo version, not all communities included in the original dataset are available, since most of them are too large to be interactively displayed in-browser on commonly available computer systems. However, the visualised features were computed for all communities for possible downstream analysis applications.

\begin{figure*}
\centering
\includegraphics[width=\textwidth]{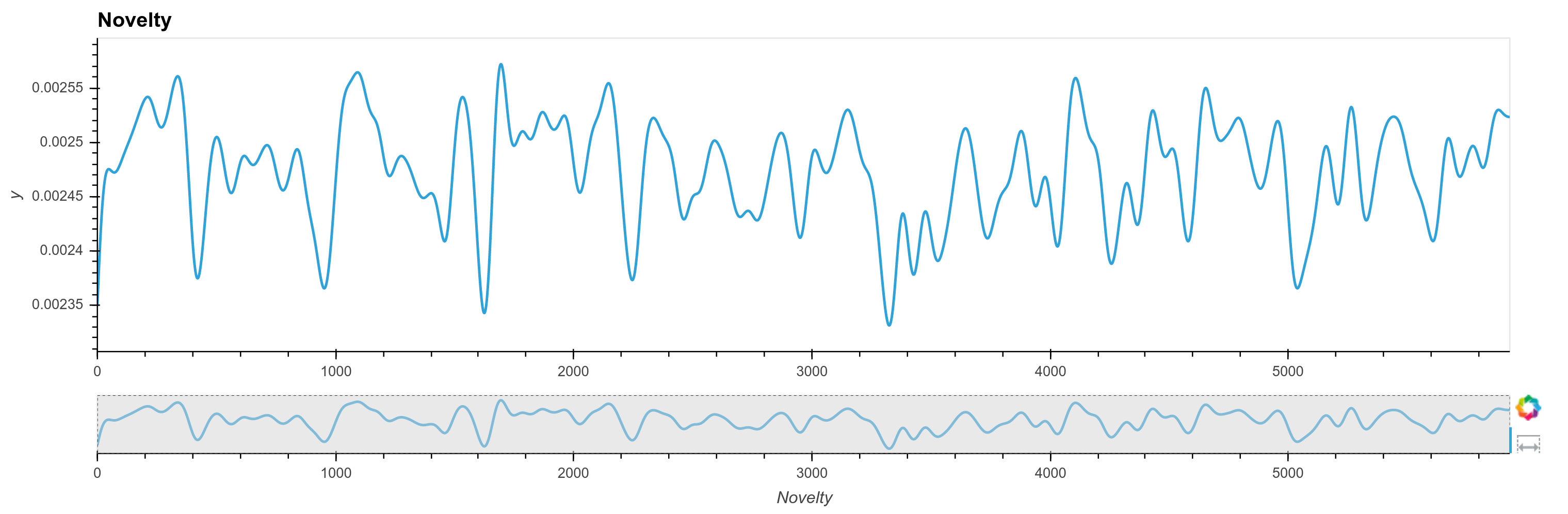}
\caption{Example plot for novelty over time}
\label{fig:novelty}
\end{figure*}

\printbibliography

@article{cer:2018,
  author    = {Daniel Cer and
               Yinfei Yang and
               Sheng{-}yi Kong and
               Nan Hua and
               Nicole Limtiaco and
               Rhomni St. John and
               Noah Constant and
               Mario Guajardo{-}Cespedes and
               Steve Yuan and
               Chris Tar and
               Yun{-}Hsuan Sung and
               Brian Strope and
               Ray Kurzweil},
  title     = {Universal Sentence Encoder},
  journal   = {CoRR},
  volume    = {abs/1803.11175},
  year      = {2018}
}

@article {barron:2018,
	author = {Barron, Alexander T. J. and Huang, Jenny and Spang, Rebecca L. and DeDeo, Simon},
	title = {Individuals, institutions, and innovation in the debates of the French Revolution},
	volume = {115},
	number = {18},
	pages = {4607--4612},
	year = {2018},
	doi = {10.1073/pnas.1717729115},
	publisher = {National Academy of Sciences},
	issn = {0027-8424}
}

@inproceedings{sun:2020,
  author    = {Zhen Sun and
               Roei Schuster and
               Vitaly Shmatikov},
  editor    = {Hugo Larochelle and
               Marc'Aurelio Ranzato and
               Raia Hadsell and
               Maria{-}Florina Balcan and
               Hsuan{-}Tien Lin},
  title     = {De-Anonymizing Text by Fingerprinting Language Generation},
  booktitle = {Advances in Neural Information Processing Systems 33: Annual Conference
               on Neural Information Processing Systems 2020, NeurIPS 2020, December
               6-12, 2020, virtual},
  year      = {2020},
}

@article{mcinnes:2018,
  author    = {Leland McInnes and
               John Healy},
  title     = {{UMAP:} Uniform Manifold Approximation and Projection for Dimension
               Reduction},
  journal   = {CoRR},
  volume    = {abs/1802.03426},
  year      = {2018},
  archivePrefix = {arXiv},
  eprint    = {1802.03426},
}

@article{nolet:2020,
  author    = {Corey J. Nolet and
               Victor Lafargue and
               Edward Raff and
               Thejaswi Nanditale and
               Tim Oates and
               John Zedlewski and
               Joshua Patterson},
  title     = {Bringing {UMAP} Closer to the Speed of Light with {GPU} Acceleration},
  journal   = {CoRR},
  volume    = {abs/2008.00325},
  year      = {2020},
  archivePrefix = {arXiv},
  eprint    = {2008.00325},
}

@inproceedings{hinton:2002,
  author    = {Geoffrey E. Hinton and
               Sam T. Roweis},
  editor    = {Suzanna Becker and
               Sebastian Thrun and
               Klaus Obermayer},
  title     = {Stochastic Neighbor Embedding},
  booktitle = {Advances in Neural Information Processing Systems 15 [Neural Information
               Processing Systems, {NIPS} 2002, December 9-14, 2002, Vancouver, British
               Columbia, Canada]},
  pages     = {833--840},
  publisher = {{MIT} Press},
  year      = {2002},
}

@misc{gwern:2015,
    author = {Gwern Branwen and Nicolas Christin and David Décary-Hétu and
              Rasmus Munksgaard Andersen and StExo and El Presidente and Anonymous
              and Daryl Lau and Sohhlz, Delyan Kratunov and Vince Cakic and Van Buskirk
              and Whom and Michael McKenna and Sigi Goode},
title = {Dark Net Market archives, 2011-2015},
howpublished=  {\url{https://www.gwern.net/DNM-archives}},
url = {https://www.gwern.net/DNM-archives},
type = {dataset},
year = {2015},
month = {07},
note = {Accessed: 2021-02-11} }

@inproceedings{ester:1996,
  author    = {Martin Ester and
               Hans{-}Peter Kriegel and
               J{\"{o}}rg Sander and
               Xiaowei Xu},
  editor    = {Evangelos Simoudis and
               Jiawei Han and
               Usama M. Fayyad},
  title     = {A Density-Based Algorithm for Discovering Clusters in Large Spatial
               Databases with Noise},
  booktitle = {Proceedings of the Second International Conference on Knowledge Discovery
               and Data Mining (KDD-96), Portland, Oregon, {USA}},
  pages     = {226--231},
  publisher = {{AAAI} Press},
  year      = {1996},
}

@InProceedings{stein:2019h,
  author =              {Janek Bevendorff and Benno Stein and Matthias Hagen and Martin Potthast},
  booktitle =           {14th Conference of the North American Chapter of the Association for Computational Linguistics: Human Language Technologies (NAACL 2019)},
  editor =              {Jill Burstein and Christy Doran and Thamar Solorio},
  month =               jun,
  pages =               {654-659},
  publisher =           {{ACL}},
  site =                {Minneapolis},
  title =               {{Generalizing Unmasking for Short Texts}},
  url =                 {https://www.aclweb.org/anthology/N19-1068},
  year =                2019
}

@inproceedings{koppel:2004,
  author    = {Moshe Koppel and
               Jonathan Schler},
  editor    = {Carla E. Brodley},
  title     = {Authorship verification as a one-class classification problem},
  booktitle = {Machine Learning, Proceedings of {(ICML} 2004},
  series    = {{ACM} International Conference Proceeding Series},
  volume    = {69},
  publisher = {{ACM}},
  year      = {2004},
  doi       = {10.1145/1015330.1015448},
}

@inproceedings{wolf:2020,
    title = "Transformers: State-of-the-Art Natural Language Processing",
    author = "Thomas Wolf and Lysandre Debut and Victor Sanh and Julien Chaumond and Clement Delangue and Anthony Moi and Pierric Cistac and Tim Rault and Rémi Louf and Morgan Funtowicz and Joe Davison and Sam Shleifer and Patrick von Platen and Clara Ma and Yacine Jernite and Julien Plu and Canwen Xu and Teven Le Scao and Sylvain Gugger and Mariama Drame and Quentin Lhoest and Alexander M. Rush",
    booktitle = "Proceedings of the 2020 Conference on Empirical Methods in Natural Language Processing: System Demonstrations",
    month = oct,
    year = "2020",
    address = "Online",
    publisher = "ACL",
    pages = "38--45"
}

@article{radford:2019,
  title={Language Models are Unsupervised Multitask Learners},
  author={Radford, Alec and Wu, Jeff and Child, Rewon and Luan, David and Amodei, Dario and Sutskever, Ilya},
  year={2019}
}

@misc{alsayra:2011,
    author={{Alsayra (web forum)}}, 
    url={http://azsecure-data.org/other-forums.html}, 
    journal={AZSecure - Data for Intelligence and Security Informatics},
    year={2011--2012}
}

@inproceedings{devlin:2019,
  author    = {Jacob Devlin and
               Ming{-}Wei Chang and
               Kenton Lee and
               Kristina Toutanova},
  editor    = {Jill Burstein and
               Christy Doran and
               Thamar Solorio},
  title     = {{BERT:} Pre-training of Deep Bidirectional Transformers for Language
               Understanding},
  booktitle = {Proceedings of {NAACL-HLT} 2019},
  pages     = {4171--4186},
  publisher = {Association for Computational Linguistics},
  year      = {2019},
  doi       = {10.18653/v1/n19-1423},
}

\end{document}